\documentclass[review]{elsarticle}

\usepackage{amssymb}
\graphicspath{{./Images/}}
\usepackage{color}
\usepackage{amsmath}
\usepackage{bm}
\usepackage{soul}

\usepackage{multirow}
\usepackage{diagbox}
\usepackage{rotating}
\usepackage{booktabs}
\usepackage{colortbl}
\usepackage{overpic}
\usepackage[table]{xcolor}
\usepackage{textcomp}
\usepackage{contour}
\usepackage{textcomp}

\usepackage{subfig}
\usepackage{pdfpages}
\usepackage{lscape}
\usepackage{makecell}
\usepackage{setspace}

\usepackage{algorithmic}

\usepackage[ruled]{algorithm2e}
\usepackage{amssymb,amsmath,caption}
\usepackage[hang,flushmargin]{footmisc}
\usepackage{lipsum}
\makeatletter

\SetKwInput{KwInput}{Input}
\SetKwInput{KwOutput}{Output}

\newcommand{\algorithmfootnote}[2][\footnotesize]{%
	\let\old@algocf@finish\@algocf@finish
	\def\@algocf@finish{\old@algocf@finish
		\leavevmode\rlap{\begin{minipage}{\linewidth}
				#1#2
		\end{minipage}}%
	}%
}

\definecolor{mygreen}{RGB}{0,150,0}
\definecolor{myred}{RGB}{200,0,0}
\usepackage[pagebackref=false,letterpaper=true,colorlinks=true,linkcolor=myred,citecolor=blue,bookmarks=false]{hyperref}

\journal{Neurocomputing}

\newcommand{\figref}[1]{Fig. \ref{#1}}
\newcommand{\tabref}[1]{Tab. \ref{#1}}

\newcommand{\algref}[1]{Alg. \ref{#1}}

\def\ie{\emph{i.e.}}
\def\eg{\emph{e.g.}}

\def\etal{\textit{et al.~}}
\def\ourmodel{\textit{MCI-Net}}

\begin{document}

	\begin{frontmatter}
		
		
		
		\title{Multi-level Cross-modal Interaction Network for RGB-D Salient Object Detection}
		
		\author[uestc]{Zhou Huang}
		\ead{chowhuang23@gmail.com}
		\author[uestc]{Huai-Xin Chen \corref{cor1}}
		\ead{huaixinchen@uesct.edu.cn}
		\author[NUST]{Tao Zhou}
		\ead{taozhou.ai@gmail.com}
		\author[CETC]{Yun-Zhi Yang}
		\ead{yangyz@cetca.net.cn}
		\author[uestc]{Bi-Yuan Liu} 
		\ead{byliu@std.uestc.edu.cn}  
		\address[uestc]{University of Electronic Science and Technology of China, Chengdu, China}
		\address[NUST]{Nanjing University of Science and Technology, Nanjing, China}
		\address[CETC]{CETC Special Mission Aircraft System Engineering Co.,Ltd, Chengdu, China}
		
		\cortext[cor1]{Corresponding author.}
		
		
		\begin{abstract}
			Depth cues with affluent spatial information have been proven beneficial in boosting salient object detection (SOD), while the depth quality directly affects the subsequent SOD performance. However, it is inevitable to obtain some low-quality depth cues due to limitations of its acquisition devices, which can inhibit the SOD performance. Besides, existing methods tend to combine RGB images and depth cues in a direct fusion or a simple fusion module, which makes they can not effectively exploit the complex correlations between the two sources. Moreover, few methods design an appropriate module to fully fuse multi-level features, resulting in cross-level feature interaction insufficient. To address these issues, we propose a novel Multi-level Cross-modal Interaction Network (\ourmodel) for RGB-D based SOD. Our \ourmodel~ includes two key components: 1) a cross-modal feature learning network, which is used to learn the high-level features for the RGB images and depth cues, effectively enabling the correlations between the two sources to be exploited; and 2) a multi-level interactive integration network, which integrates multi-level cross-modal features to boost the SOD performance. Extensive experiments on six benchmark datasets demonstrate the superiority of our \ourmodel~ over 14 state-of-the-art methods, and validate the effectiveness of different components in our \ourmodel~. More important, our \ourmodel~ significantly improves the SOD performance as well as has a higher FPS.
			
		\end{abstract}
		
		
		%
		%
		%
		%
		%
		
		
	\end{frontmatter}
	
	
	\section{Introduction}
	Salient object detection (SOD) aims at automatically identifying salient regions in a scene from their surroundings~\cite{,fan2020taking,zhou2021rgb,huang2020contrast}, which has drawn increasing interest in computer vision. As a pre-processing tool, SOD benefits several real-world applications, including object segmentation \cite{wang2017saliency}, object tracking \cite{mahadevan2012biologically}, image enhancement \cite{kim2006saliency}, person re-identification \cite{zhao2016person}, and so on. Although many SOD methods have been developed and obtained good results over the past several years, they still face several limitations, especially when subject to varying illuminations and complex backgrounds. Recently, with the surge in popularity of depth sensors in smart devices, depth cues have been used to provide complementary shape and spatial layout information to overcome these challenges. Consequently, determining how to effectively fuse RGB images and depth cues to improve the SOD performance is a critical problem in dealing with RGB-D data.

	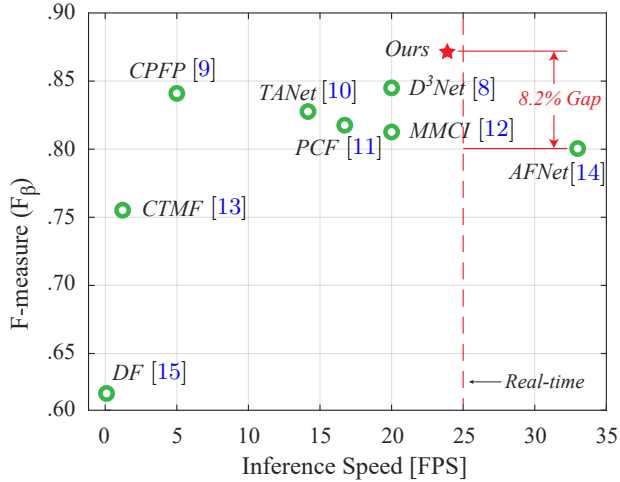
\begin{figure}[t]
		\centering
		\begin{overpic}[width=0.7\linewidth]{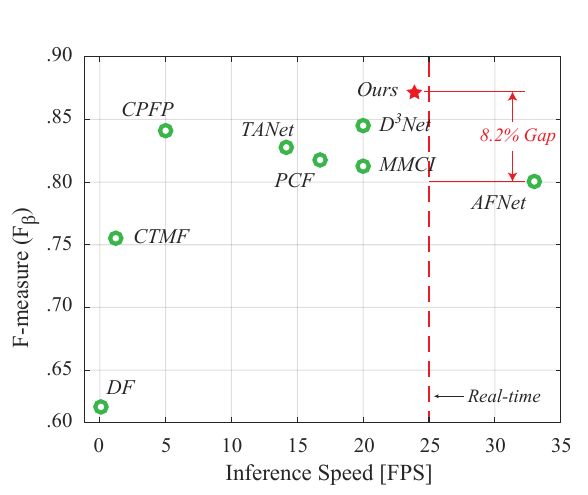}
			\put(74.3,61.8){\small \cite{fan2019rethinking}}
			\put(30.5,64.7){\small \cite{zhao2019contrast}}
			\put(50.7,61.2){\small \cite{chen2019three}}
			\put(54.6,52.5){\small \cite{chen2018progressively}}
			\put(75.3,55.3){\small \cite{chen2019multi}}
			\put(33.4,43){\small \cite{han2017cnns}}
			\put(89.6,48.4){\small \cite{wang2019adaptive}}
			\put(24,17.2){\small \cite{qu2017rgbd}}
			
		\end{overpic}
		\caption{F-measure vs. Inference Speed (\ie, FPS) on the STERE dataset \cite{niu2012leveraging}. Our model achieves comparable accuracy compared to the state-of-the-art methods, including D$^3$Net \cite{fan2019rethinking}, CPFP \cite{zhao2019contrast}, TANet \cite{chen2019three}, PCF \cite{chen2018progressively}, DF \cite{qu2017rgbd}, MMCI \cite{chen2019multi}, CTMF \cite{han2017cnns}, and AFNet \cite{wang2019adaptive}, at a significantly higher FPS.}\label{fig:model0}
	\end{figure}

	Given paired RGB and depth images, several methods have been developed for RGB-D based SOD. Early works \cite{desingh2013depth,cheng2014depth} mainly focused on hand-crafted low-level features. Further, the early fusion strategy involved either directly cascading RGB-D channels, or combining the decision maps obtained from each modality-independent output. In these simple fusion methods, the correlations between cross-modal data are ignored. Recently, due to the success of CNN in learning powerful feature representations, various works have employed these to achieve SOD from RGB-D data. For example, Wang \etal \cite{wang2019adaptive} proposed an adaptive fusion network, in which an adaptive learning switch map is designed to integrate effective information from RGB and depth predictions, and an edge-preserving loss is used for correcting blurry boundaries. Zhu \etal \cite{zhu2019pdnet} proposed a prior-model guided master network to process RGB information, which was pre-trained on a conventional RGB dataset to overcome  the shortage of training data. In order to enhance the RGB-D representational ability and achieve selective cross-modal fusion, Chen \etal \cite{chen2019three} proposed a three-stream architecture with an attention-aware cross-modal fusion network. Overall, these RGB-D SOD methods based on deep models have achieved significant improvements over hand-crafted features based approaches. In addition, in some works \cite{han2017cnns,chen2018progressively,liu2019salient}, the original depth map is directly encoded by HHA (\emph{i.e.} horizontal disparity, height above ground, and the angle of the local surface normal with the inferred direction of gravity.), and such a processing method can improve the quality of the input source to a certain extent \cite{Gupta2014Learning,chen2019three,li2020icnet}.

	Although great progress has been made in this field, existing RGB-D based SOD methods still face several issues. \emph{First}, the quality of the captured depth cues varies tremendously across different conditions, which can inhibit the SOD performance. \emph{Second}, most existing methods combine RGB images and depth cues using either an early fusion or late fusion strategy, however, this direct combination operation or a designed simple fusion module can not effectively exploit the complex correlations between the two sources. While some works have introduced a middle-fusion or multi-scale fusion \cite{zhao2019contrast,chen2018progressively,han2017cnns,shigematsu2017learning}, it is still challenging to design an appropriate and effective module for exploring the multi-level interactive information.


	To this end, we propose a novel Multi-level Cross-modal Interaction Network (\ourmodel) for RGB-D salient object detection, which consists of two key components, \ie, a cross-modal feature learning network and multi-level interactive integration network. Specifically, in the cross-modal feature learning network, the depth cues encoded as enhanced HHA are first used for cross-modal feature learning in a two-stream structure module, while a cross-modal refinement module (CMRM) is proposed to integrate cross-modal features. In the multi-level interactive integration network, a multi-level fusion module (MLFM) is developed to fuse the features of each level in a bottom-up manner. In addition, a feedback integration module (FIM) is proposed to propagate the features of the last convolutional layer back to the previous layers. Finally, we fully integrate the cross-modal features from different levels in a pyramid style. Extensive experiments on six benchmark datasets demonstrate the effectiveness of the proposed \ourmodel~ against 14 state-of-the-art (SOTA) methods in terms of evaluation metrics,  and the extended experiments further demonstrate the compatibility and robustness of the model. Moreover, 
		when comparing with several current SOTA methods (including D$^3$Net \cite{fan2019rethinking}, CPFP \cite{zhao2019contrast}, TANet \cite{chen2019three}, PCF \cite{chen2018progressively}, DF \cite{qu2017rgbd}, MMCI \cite{chen2019multi}, CTMF \cite{han2017cnns}, and AFNet \cite{wang2019adaptive}), our \ourmodel~ significantly improves the SOD performance and also has a higher FPS, as shown in Fig.~\ref{fig:model0}. The main \textbf{contributions} of this work are as follows:\vspace{-0.1cm}
	\begin{itemize}
		
		\item [(1)] 
		We propose a novel \ourmodel~ for RGB-D salient object detection, which can effectively exploit the correlation between RGB images and depth cues, while also exploring multi-level information to boost the SOD performance.
		
		\item [(2)] 
		To exploit the correlations across RGB images and depth cues, a CMRM is proposed to integrate cross-modal features. The CMRM is carried out in various of levels of feature spaces for further multi-level feature fusion.
		

		\item [(3)] 
		A multi-level interactive integration network is proposed to fully integrate the cross-modal features from different levels, enabling multi-level interactive information to be explored. An MLFM is proposed to fuse the features of each level in a bottom-up manner, and an FIM is proposed to propagate the features of the last convolutional layer back to the previous layers to reduce the information lost during downsampling as well as the effect of noise.

		\item [(4)] 
		Extensive experimental results on six RGB-D SOD benchmark datasets demonstrate the effectiveness of our \ourmodel~ over other SOTA methods. Besides, we conduct a comprehensive ablation study to validate the effectiveness of different components in our \ourmodel.
		
	\end{itemize}

	\section{Related Work}
	
	\subsection{RGB-D Saliency Detection}
	
	\subparagraph{\textbf{Traditional Methods.}} According to the stage of fusion, traditional methods can be summarized into three categories: 1) input fusion, 2) feature fusion, 3) result fusion. For input fusion, Peng \etal \cite{peng2014rgbd} serialized RGB and corresponding depth cues into four channels simultaneously as a multi-stage saliency detection model. For the second category, Ju \etal \cite{ju2014depth} considered both fine-grained global structures and coarse-grained local details, and proposed to use anisotropic center differences to measure the significance of depth cues. Based on the observation that objects surrounding the background in an angular direction present a unique structure and have high saliency, Feng \etal \cite{feng2016local} proposed an RGB-D saliency feature captured by the local background enclosure (LBE). For the last category, multiple prediction results are integrated into separate post-processing steps. For example, Desingh \etal \cite{desingh2013depth} used nonlinear support vector regression to fuse multiple saliency prediction maps. In \cite{cong2016saliency}, a RGBD saliency model was proposed by combining depth confidence analysis and multiple cues fusion. 

	\subparagraph{\textbf{Deep Learning based Methods.}} Recently, convolutional neural networks (CNN) have been widely used in RGB-D saliency detection. As a pioneering work, Qu \etal \cite{qu2017rgbd} developed a method to fuse different low-level saliency cues into hierarchical features using the CNN framework, in order to effectively locate salient regions from RGB-D images. In a more recent work, Zhao \etal \cite{zhao2019contrast} integrated enhanced depth cues with RGB features through a fluid pyramid for SOD. In order to fill the gaps of SOD in real human activity scenes, Fan \etal \cite{fan2019rethinking} proposed a simple baseline architecture through depth depurator units and a feature learning module, and obtained satisfactory results. In order to treat information from different sources discriminatively and capture the continuity of cross-modal features, Li \etal \cite{li2020icnet} proposed an information conversion network using the Siamese structure with an encoder-decoder architecture. In addition, some authors have also adopted a joint learning strategy~\cite{Fu2020JLDCF}, bilateral attention~\cite{Zhang2020BANet}, and conditional variational autoencoders (CVAE)~\cite{Zhang2020UCNet} to address this task.

	\subsection{Multi-level Feature Integration}
	
	For input image pairs, some works have been devoted to studying the integration of multi-level features \cite{han2017cnns,shigematsu2017learning,chen2018progressively,zhao2019contrast}. 
	For example, Han \etal \cite{han2017cnns} proposed a multi-view CNN fusion model through a combination layer connecting the representation layers of the RGB and depth data to effectively integrate the two domains. In \cite{shigematsu2017learning}, the depth features were combined by concatenating manually designed depth features and low-level and high-level RGB saliency features. Chen \etal \cite{chen2018progressively} designed an architecture based on complementarity-aware fusion (CA-Fuse) module by cascading the CA-Fuse module and adding level-wise supervision from deep to shallow layers. Chen \etal \cite{chen2019multi} proposed a multi-scale multi-path network with cross-module interaction to enable sufficient and efficient fusion. 
	Most of these methods utilize a direct combination operation or a simple fusion module, however, the complex correlations between the RGB images and depth cues can be not effectively exploited. Further, no appropriate module has been designed for exploring the multi-level interactive information. 
	Recently, in \cite{piao2019depth}, residual connections were used to design a depth refinement block for fusing multi-level paired complementary cues from RGB and depth streams. Li \etal \cite{li2020asif} proposed an attention-steered interweave fusion network (ASIF-Net), which gradually integrates the features of the RGB image and corresponding depth map under the control of an attention mechanism. Liu \etal \cite{liu2020cross} proposed a cross-modal adaptive gated fusion generative adversarial network, which progressively combines the deep semantic features processed by the depth-wise separable residual convolution module with the side-output features of the encoder network.
	
	\section{Proposed Method}\label{sec:method}
	In this section, we first introduce the overall architecture of the proposed \ourmodel~ in Sec. \ref{sec:3.1}. Then we describe the cross-modal feature learning network and multi-level interactive integration network in Sec. \ref{sec:3.2} and Sec. \ref{sec:3.3}, respectively. Finally, the loss function is given in Sec. \ref{sec:3.4}.


	\subsection{Overall Architecture}\label{sec:3.1}
	\figref{fig:model} shows the overall architecture of the proposed \ourmodel, consisting of two key components: the cross-modal feature learning network and multi-level interactive integration network. Specifically, in the cross-modal feature learning network, each stream of the two-stream structure is built using the same construction as the public backbone network (\emph{e.g.} VGGNet \cite{Karen2015VGG}, ResNet \cite{he2016deep}, Res2Net \cite{gao2019res2net}. For an input image with a size of M$\times$N, we use backbone network to extract its features at five different levels, denoted as $ \left\lbrace f_i|i=1, \dots, 5 \right\rbrace $ with resolutions $ \left[ M/2^{i-1}, N/2^{i-1} \right]  $. Due to the high computational overhead and low performance \cite{wu2019cascaded} when using low-level features, here we only use high-level features from the last four layers with low resolutions (\ie, $ f_2 $, $ f_3 $, $ f_4 $, and $ f_5 $). In the multi-level interactive integration network, we adopt the proposed CMRM to achieve feature integration for the paired side-output features. In order to fully integrate and make use of the features of each level, we utilize the proposed MLFM to fuse them in a bottom-up manner, while fusing multi-level cross-modal features in a pyramid style. Moreover, we propose a feedback prediction module to propagate the features of the last convolutional layer back to the previous layers. These details will be given in the following sections.

	\subsection{Cross-modal Feature Learning}\label{sec:3.2}
\begin{figure*}[t]
		\centering
		\begin{overpic}[width=\textwidth]{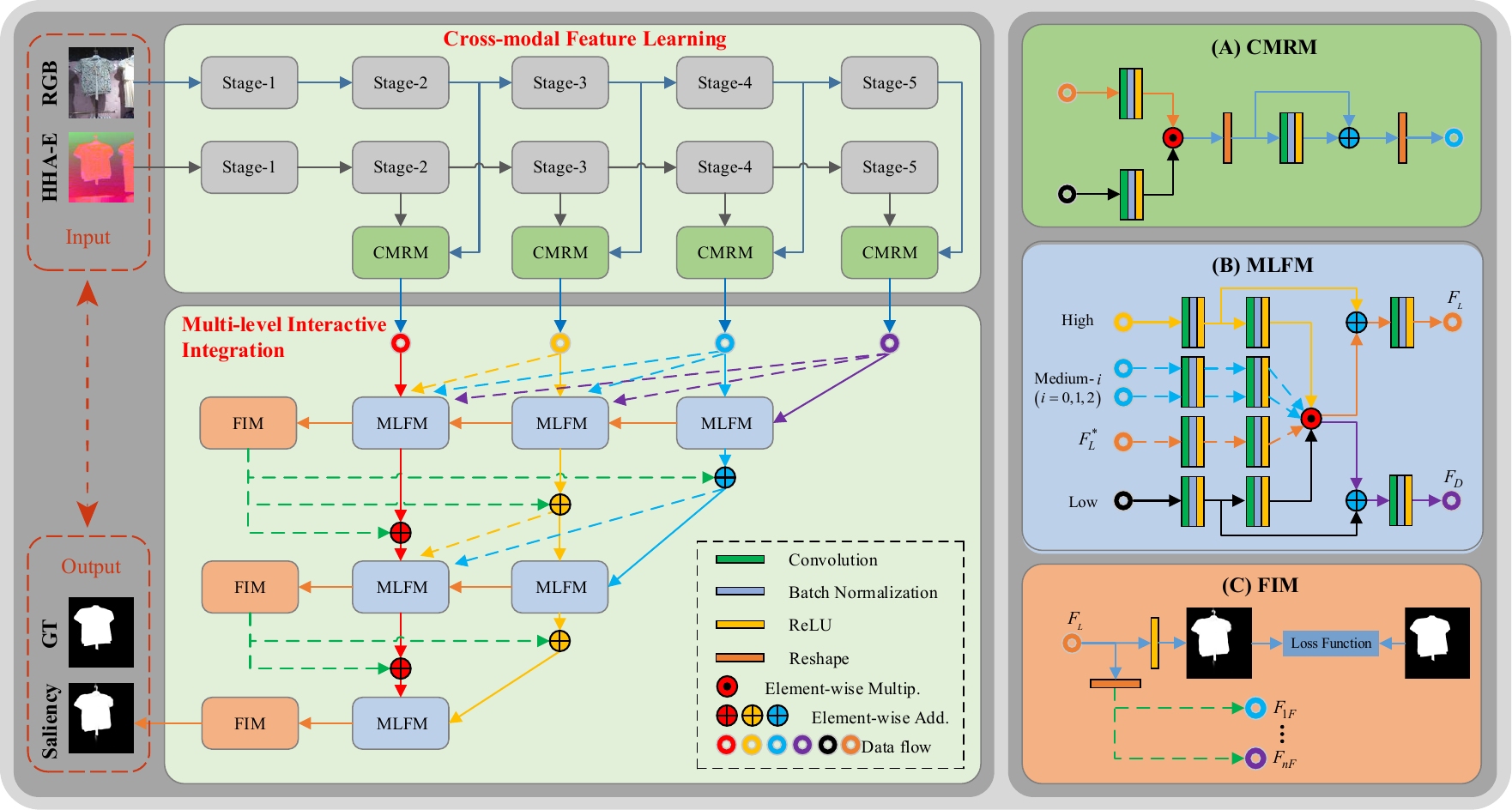}
		\end{overpic}\vspace{-0.25cm}
		\caption{
			The overall architecture of the proposed \ourmodel. Our $\ourmodel$ consists of two key components: 1) a cross-modal feature learning network, in which each stream of the two-stream structure is used to learn high-level features for each modality of data (RGB or depth images), while a CMRM is proposed to fuse cross-modal features; 2) a multi-level interactive integration network, in which we use an MLFM to fuse the features of each level in a bottom-up manner, and formulate multi-level cross-modal feature integration in a pyramid style. Besides, a feedback prediction module is used to propagate the features of the last convolutional layer back to the previous layers.}\label{fig:model}
	\end{figure*}

	The cross-modal feature learning network (as shown in \figref{fig:model}) is used to learn high-level features for the RGB images and depth cues, and exploits the correlations between the two sources. We provide details of the proposed depth map enhancement method and cross-modal refinement module below.

	\begin{figure}[t]
		\centering
		\begin{overpic}[width=\linewidth]{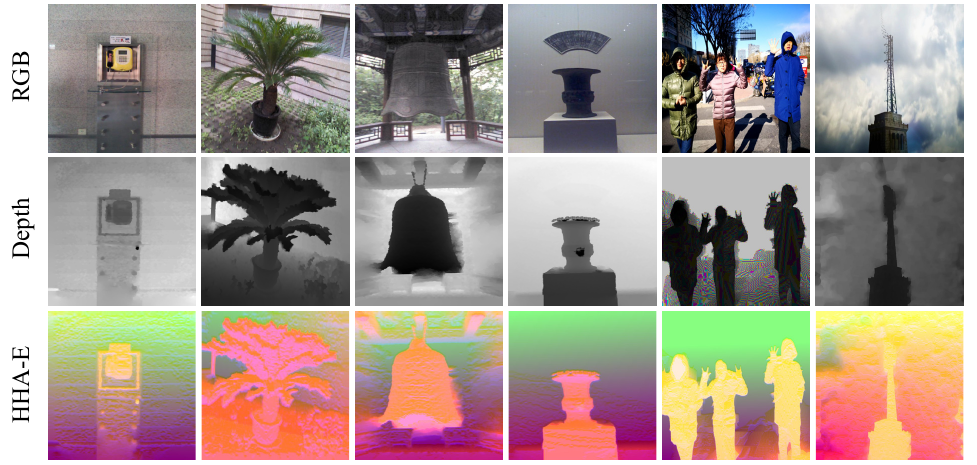}
		\end{overpic}\vspace{-0.3cm}
		\caption{Visual comparison between depth images and their enhanced maps.}
		\label{fig:hha-e}\vspace{-0.3cm}
	\end{figure}
	
	\textbf{Depth Map Enhancement}. In practice, depth maps often suffer from noise, blurred edges, and low contrast, reducing the final SOD performance when directly used for feature fusion. To overcome this, we propose a novel depth map enhancement method. A visual comparison between depth images and their enhanced maps is shown in \figref{fig:hha-e}. Specifically, considering the uneven distribution of gray values in the depth maps, we use the Otsu algorithm \cite{Otsu2007A} to obtain the threshold $T^{*} $ of the depth map for enhancing its contrast. Besides, due to the lack of a pre-trained model suitable for single-channel input when using depth maps, we encode the enhanced depth maps as a three-channel HHA with more geometric information \cite{Gupta2014Learning}. Consequently, the enhanced depth maps (\ie, HHA-E) can be directly fed into the pretrained CNN models to learn more effective feature representations for further fusion. The proposed depth map enhancement method can be defined as follows:
	\begin{equation}\label{equ:1}
		HHA\small{-}E\gets HHA\left\lbrace 
		\begin{aligned}
			\lambda_1\left( I_d<T^{*}\right) \\
			\lambda_2\left( I_d\ge T^{*}\right) \\
		\end{aligned}
		\right. ,
	\end{equation}
	where $I_d$ is the grayscale intensity, $ T^{*}=\mathop{\arg\max}\limits_{t}( \delta_{I_d<t}^2-\delta_{I_d\ge t}^2)  $ with $ t\in \left\lbrace 0,1,\dots,255\right\rbrace  $, and $ \lambda_1 $ and $\lambda_2 $ are scaling parameters.

	\textbf{Cross-modal Refinement Module}. Considering that the depth and RGB cues collected from different sources are strongly complementary, we design an effective cross-modal refinement module (CMRM) to fully extract and fuse paired cross-modal features. Because multiple input sources have the same number of channels and rich features, and also have the same processing in the backbone network. In order to learn the input feature residuals, unlike the DRB module proposed by Piao et al. \cite{piao2019depth}, we process $ S_i^{RGB}$ and $ S_i^{HHA-E}$ through a combined set of weight layers and then modulate the features by element multiplication. As shown in \figref{fig:model} (A), in the input of CMRM, $ S_i^{RGB}$ and $ S_i^{HHA-E}~(i\in\left[2,5\right])$ represent the side-out features of the $i$-th level from the RGB and enhanced depth streams, respectively. First, we feed $ S_i^{RGB}$ and $ S_i^{HHA-E}$ into a series of weight layers $\mathcal{W}$ containing a convolutional layer with 3$\times$3 kernels, a batch normalization (BN) layer and a ReLu activation layer to learn a depth residual. Then, cross-modal features are modulated by element-wise multiplication to refine the required feature parts. Next bilinear interpolation or maximum pooling is used to reshape each level of the fusion feature to be the same resolution. To further enhance the fused features, we integrate the features processed by $\mathcal{W}$ to the previously fused features through a residual connection (\emph{i.e.}, element-wise summation), and then reshape the channel size to obtain the final enhanced fusion feature. The process of CMRM can be given as:
	
	\begin{equation}\label{equ:2}
		\left\{
		\begin{aligned}
			&F_{fuse}^i=\mathcal{R}\left( \mathcal{W}\left(S_i^{RGB}\right)\odot\mathcal{W}\left(S_i^{HHA-E}\right) \right)\\
			&F_{\ast}^i=\mathcal{R}\left(F_{fuse}^i\oplus\mathcal{W} \left(F_{fuse}^i \right) \right),
		\end{aligned} 
		\right.
	\end{equation}
	where $\odot$ and $ \oplus$ denote element-wise multiplication and addition, respectively, and $ \mathcal{R} $ denotes a reshaping operation. By applying CMRM at each level, the module effectively fuses and learns discriminative depth and RGB features, generating four cross-modal refinement features from different levels. It is worth noting that the refined fusion strategy combines local spatial detail information and global semantic information to improve model performance.
	
	\subsection{Multi-level Interactive Integration}\label{sec:3.3}
	
	The multi-level interactive integration network (as shown in \figref{fig:model}) is proposed to explores and fuse  multi-level interactive information to boost the SOD performance. In this integration network, we propose a multi-level fusion module (MLFM) to effectively integrate multiple cross-level features, and a feedback integration module (FIM) to propagate the features of the last layer back to previous layers. The details of the two modules are provided below.

	\textbf{Multi-level Fusion Module}. The multi-level features obtained from the previous stage are statistically different. Although high-level features may lose a lot of detailed information after a series of downsampling processes, they still have highly consistent semantic information and clear background. On the other hand, due to the limitations of the receiving field, the low-level features retain rich detailed information and noise, which is critical for generating a saliency map with a clear outline. To take advantage of the benefits and alleviate the issues of high- and low-level features, we propose a multi-level fusion module (MLFM) to effectively integrate multiple cross-level features, as shown in \figref{fig:model} (B) (Note that the medium features represented by dotted lines in the figure may not be available in some input features.). Unlike the fusion module proposed by Wei et al. \cite{wei2020f3net}, our CLFM integrates multiple levels of features and passes the features back through interactive strategies. Specifically, the proposed MLFM includes two stages. Firstly, a convolutional layer with a 3$\times$3 kernel size is applied to multiple cross-level features (low-level features $F_l$, medium-level features $F_m $, and high-level features $F_h $) to adapt to subsequent processing, and cross-level features are integrated by element-wise multiplication to retain their common parts. Second, the fused features $F_{fused} $ are combined with the original low-level features and high-level features by element-wise summation and used as the output $F^{*} $ of the module. The above process can be formulated as:
	\begin{equation}\label{equ:3}
		\left\{
		\begin{aligned}
			&F_{fused}= \mathcal{W}^2\left(F_l\right)\odot\mathcal{W}^2(F_m^{n})\odot\mathcal{W}^2\left(F_h \right), ~n\in\left\lbrace 0,2,3 \right\rbrace \\
			&F_{x}^{*}=\mathcal{W}^1\left( \mathcal{W}^1\left(F_{x} \right)\oplus F_{fused}\right), ~x\in\left\lbrace l,h \right\rbrace
		\end{aligned} 
		\right.
	\end{equation}
	where $\mathcal{W}^1$ represents a single operation including a combination of convolutional, BN and Relu layers, and $\mathcal{W}^2$ means twice. By processing a series of MLFM units, multi-level features gradually complement each other with effective information; that is, the contours of high-level features are sharpened, and the backgrounds of low-level features become more consistent.

	\begin{algorithm}[t]\small
		\caption{Multi-level interactive integration}
		\label{alg:1}
		\algorithmfootnote{Note: $L$ \& $D$ are the flow of data left and down, $F$ is feedback.}
		\KwInput{Cross-level features: $ \left\lbrace F_i^P|i\in\left[5,2 \right]  \right\rbrace $, \\\hspace{0.855cm}Number of pyramid layers: $P=4$.}
		\For{$ m=P:2 $}
		{
			\For{$ n=m:2 $}
			{
				\eIf{$F_{nL}^m=null$}
				{
					$ \left\lbrace F_{nL}^m,F_{nD}^{m-1}\right\rbrace \gets MLFM\left(\left\lbrace F_i^m|i=\left[ m+1,n\right] \right\rbrace \right) $;
				}
				{
					$ \left\lbrace F_{nL}^m,F_{nD}^{m-1}\right\rbrace \gets MLFM\left(\left\lbrace F_i^m|i=\left[ m+1,n\right] \right\rbrace,F_{nL}^m\right) $;
				}
			}
			$ \left\lbrace F_{iF}^{m-1}|i\in\left[m,2 \right] \right\rbrace\gets FIM\left( F_{2L}^m\right) $;\\
			
			\If{$m\geq3$}
			{
				$ \left\lbrace F_i^{m-1}|i\in \left[m,2 \right] \right\rbrace \gets \left\lbrace F_{iF}^{m-1}\oplus F_{iD}^{m-1}|i\in \left[m,2 \right] \right\rbrace  $.
			}
		}
		\KwOutput{$ Sal \gets Conv\left(F_{2L}^2 \right) $.} 
	\end{algorithm}

	\textbf{Feedback Integration Module}. When dealing with cross-modal and multi-level features, the key is to maintain the stability and compatibility of the features. Inspired by the recent multi-scale feature fusion \cite{liu2019salient,li2020icnet,zhao2019contrast}, we develop a feedback pyramid feature fusion structure with MLFM as the processing unit, as shown in \figref{fig:model} (C). This structure introduces the high-level (low-resolution) features of each layer into the low-level (high-resolution) features through a pyramid connection to make full use of the features at multiple levels, which helps maintain the stability and compatibility of the learned features. Specifically, the designed pyramid structure includes two main processes: from right to left and top-down. For the process from right to left, we guide and fuse high-level features to low-level features through MLFM, and then supervise the integrated features $ F_L $ in the feedback integration module (FIM) and feed back the corresponding features  $ F_F $ to the next layer of the pyramid. For the top-down process, the feature flows are cross-modal refinement features (\eg, in the first layer, $ \left\lbrace F_i^4|i\in\left[5,2 \right]  \right\rbrace  $ ) of different levels in the previous layer. We combine the features $ F_D $ obtained by the MLFM with the feedback features after downsampling, and then input them to the next layer for the same processing. Finally, for the output of the last layer in the pyramid structure, we use an appropriate convolution operation $ Conv\left( \cdot\right) $ to obtain the saliency map. The cross-modal and multi-level features can be fused in a pyramid style by continuous flow and integration. The multi-level interactive integration process is summarized in \algref{alg:1}. 
	
	\subsection{Loss Function}\label{sec:3.4}
	
	Considering that the adopted binary cross entropy (BCE) loss function ignores the perception of the overall structure of the image and the loss of foreground with small salient objects, we introduce the center-surround weighting item $\left(w_i\right)$ to alleviate this situation. The weighted loss can be expressed as:
	\begin{equation}\label{equ:4} 
		\begin{aligned}
			\mathcal{L}_{csBCE}\left( G,P\right)= -\dfrac{1}{\sum_{i=1}^{N}w_i}\sum_{i=1}^{N}&\left(\left( 1+w_i \right) \left( g_i\log\left( p_i\right)\right. \right.    \\
			&+ \left.\left. \left( 1-g_i\right) \log\left( 1-p_i\right) \right) \right), 
		\end{aligned}
	\end{equation}
	where  $ G\in\left\lbrace 0,1 \right\rbrace  $ and $ P\in\left[  0,1 \right]  $ respectively represent the ground truth and the predicted saliency map, $g_i\in G$, $p_i\in P$.  $ N $ represents the total number of image pixels. In $ \mathcal{L}_{csBCE}\left( G,P\right) $ , we assign an attention (importance) weight value to each pixel by $ w_i\in\left[  0,1 \right]  $, which is calculated by the following equation:
	\begin{equation}\label{equ:5}
		w_i=\left|\frac{\left\langle w_{A_i}G_{A_i}\right\rangle _1}{\left\langle G_{A_i} \right\rangle_0 }-G_i \right|,w_A=\frac{1}{2n_d}\left( \sin \left( \frac{\pi d}{\left\langle A\right\rangle _0} -\frac{\pi}{2}\right)+1  \right),  
	\end{equation}
	where $n_d$ represents the number of pixels whose Euclidean distance from the center of the image area $A$ is $d$. $A_i$ represents the surrounding area centered on the pixel $ i $. Following standard norms, we define $ \left\langle \cdot \right\rangle_0  $ and $ \left\langle \cdot \right\rangle_1  $ to represent the sum of the number of area pixels and the sum of the area pixel values, respectively.
	
	In order to use local and global information to generate more accurate salient object boundaries and reduce the impact of the uneven distribution, we utilize an enhanced position-aware loss $\left(\mathcal{L}_{EPA} \right)$ by introducing the weighted IoU loss $\left(\mathcal{L}_{WIoU} \right)$ \cite{wei2020f3net}. The $\mathcal{L}_{EPA}$ is defined as:
	\begin{equation}\label{equ:6}
		\begin{aligned}
			\mathcal{L}_{EPA}\left( G,P\right)=\mathcal{L}_{WIoU}\left( G,P \right) +\mathcal{L}_{csBCE}\left( G,P\right).
		\end{aligned}
	\end{equation}
	
	Further, we assign a smaller weight to the upper pyramid loss with a larger error. Finally, the overall loss function of our model is defined as:
	\begin{equation}\label{equ:7}
		\begin{aligned}
			\mathcal{L}_{total}=\sum_{i=1}^{3}\frac{1}{2^{i-1}}\mathcal{L}_{EPA}\left( G,P\right).  
		\end{aligned}
	\end{equation}

	\section{Experiments}
	\subsection{Datasets and Evaluation Metrics}
	\textbf{Datasets.} Experiments are conducted on six public RGB-D benchmark datasets: STERE \cite{niu2012leveraging}, NJU2K \cite{ju2014depth}, NLPR \cite{peng2014rgbd}, DES \cite{cheng2014depth}, SSD \cite{zhu2017three} and SIP \cite{fan2019rethinking}. Of these, STERE (contains 1000 images taken outdoors), NJU2K (contains 1985 images collected from 3D movies and the Internet,), NLPR (contains 1000 images captured from indoor and outdoor,), DES (contains images of indoor scenes,) SSD (contains 80 images with complex backgrounds captured from 3D movies) and SIP (contains 929 images with people captured outdoors), see \tabref{tab:1} for details. 
	For another DUTLF \cite{piao2019depth} dataset, because some methods use different training sets, we retrain the network for extended comparison experiments. In addition, we extend the RGB-Thermal (RGB-T) dataset VT1000 \cite{tu2019rgb} to further evaluate the compatibility and robustness of the proposed model.
	\begin{table}[t!]
		\small
		\centering
		\renewcommand{\arraystretch}{1.0}
		\renewcommand{\tabcolsep}{1.0mm} 
		
		\caption{Summary of the RGB-D datasets used.} \label{tab:1} 
		\begin{tabular*}{1\linewidth}{lllll}
			\toprule
			\textbf{Datasets} &\textbf{Year}&\textbf{Num.} & \textbf{Size} &\textbf{Outlines} \\
			\midrule
			STERE \cite{niu2012leveraging}&2012& 1000 & 512$\times$384& Outdoor scenes. \\
			NJU2K \cite{ju2014depth}&2014&1985 &$ 421^{\sim884}\times355^{\sim600} $&3D movies, the web, and photos.\\
			NLPR \cite{peng2014rgbd}&2014&1000 &  $480\times640$ or $ 640\times480$ &  Indoor and outdoor scenes.\\
			DES \cite{cheng2014depth}&2014&135&$ 640\times480$& Indoor scenes.\\
			SSD \cite{zhu2017three}&2017&80&$ 960\times1080$&Complex scenes in 3D movies\\
			SIP \cite{fan2019rethinking}&2019&929& $744\times992$ or $992\times744$ & Outdoor scene with people.\\
			
			\bottomrule	
		\end{tabular*}
	\end{table}

	\textbf{Evaluation Metrics.} To comprehensively evaluate various methods, we adopt five popular evaluation metrics, including mean absolute error $(MAE,~\mathcal{M} ) $ \cite{perazzi2012saliency}, S-measure $ (S_\alpha) $ \cite{fan2017structure}, F-measure $ (F_\beta) $ \cite{margolin2014evaluate,borji2015salient}, E-measure  $ (E_\xi)  $ \cite{fan2018enhanced}, and Precision-Recall $ (PR)$ curve. Following Fan \etal \cite{fan2019rethinking}, we use a series of fixed (0-255) thresholds to calculate the mean $F_\beta$ and mean $ E_\xi $. The details of these evaluation metrics are as follows:
	
	$\bullet$ {\emph{MAE}} ($\mathcal{M}$). We evaluate the \emph{Mean Absolute Error} (MAE) value between the saliency map $S$ and the binary ground-truth map $G$. The calculation formula is:
	\begin{equation}
		\mathcal{M}=\frac{1}{W*H}\sum_{i=1}^{W}\sum_{i=1}^{H}\left|S\left(i,j\right)-G\left(i,j\right) \right|,
	\end{equation}
	where $ W~\&~H $ are the width and height of the map. MAE estimates the similarity between the saliency map and the ground-truth map, and normalizes it to [0,1].\\
	
	$\bullet$ {\emph{S-measure ($S_{\alpha}$)}}. Considering the importance of image structural information, $S_{\alpha}$ \cite{fan2017structure} takes the structural similarity of regional perception ($S_r$) and object perception ($S_o$) as the evaluation of structural information. $S_{\alpha}$ is calculated as:
	\begin{equation}
		S_{\alpha}=\alpha * S_{o}+\left(1-\alpha\right)*S_{r},  ~~\alpha=0.5,
	\end{equation}
	where $ \alpha \in \left[ 0,1\right]$ is the balance parameter.\\
	
	$\bullet$ {\emph{F-measure ($F_{\beta}$)}}. $F_{\beta}$ is widely used to evaluate the performance of SOD. Following the work of Borji \cite{borji2015salient} and Fan \cite{fan2019rethinking} \etal, We use different fixed [0,255] thresholds to comprehensively evaluate the $F_{\beta}$ metric. This metric is calculated as follows:
	\begin{equation}
		F_{\beta}=\left(1+\beta ^2\right)\frac{P*R}{\beta ^{2}P+R} ,  ~~\beta ^2=0.3.
	\end{equation}\\
	
	$\bullet$ {\emph{E-measure ($E_{\phi}$)}}. $E_{\phi}$ \cite{fan2018enhanced} is a recently proposed enhanced alignment metric. This metric is based on cognitive vision and integrates local values of images with image-level averages to capture global statistical information and local pixel matching information. $E_{\phi}$ is calculated as:
	\begin{equation}
		E_{\phi}=\frac{1}{W*H}\sum_{i=1}^{W}\sum_{i=1}^{H}\phi_{FM}\left(i,j\right), 
	\end{equation}
	where $ \phi_{FM} $ denotes the enhanced-alignment matrix \cite{fan2018enhanced}.
	
	%
	%
	%
	%
	$\bullet$  \emph{PR curve}. The saliency map is binarized by a series of thresholds from 0 to 255 to generate a series of precise-recall pairs.
	\begin{equation}\label{equ:12}
		P:\frac{\left|S \cap G \right|}{\left|S \right|},  R: \frac{\left|S \cap G \right|}{\left|G \right|},
	\end{equation}

	\subsection{Implementation Details}
	Following \cite{chen2018progressively, chen2019three, zhao2019contrast, fan2019rethinking}, we randomly select 1400 and 650 image pairs from the NJU2K \cite{ju2014depth} and NLPR \cite{peng2014rgbd} datasets, respectively, as the training set. In supplementary experiments, following the same setting in \cite{piao2019depth}, we use 800 image pairs for training and the remaining 400 for testing in DUTLF dataset. For VT1000 \cite{tu2019rgb}, we randomly sample 600 image pairs for training, and the rest are used for testing.
	The parameters $\lambda_1 $ and $\lambda_2 $ in Eq. \eqref{equ:1} are set to 0.8 and 1.2, respectively. For data augmentation, we use horizontal flips, random cropping, and multi-scale operations to process input image pairs. A pre-trained ResNet-50 is used as the backbone network of our model. The maximum learning rate is set to 0.005, and the other modules are 0.05. The entire network uses stochastic gradient descent (SGD) for end-to-end training. We use the PyTorch toolbox to implement the proposed model. On a desktop with an Intel Xeon E5-2620 CPU, NVIDIA RTX 2070 GPU and 32GB RAM, the training takes 9.5 hours when batch size and maximum epoch both set to 32. For image pairs with an input size of 352$\times$352, the average inference time is 0.07s.

	\begin{figure*}[thp!]
		\centering
		\begin{overpic}[width=\linewidth]{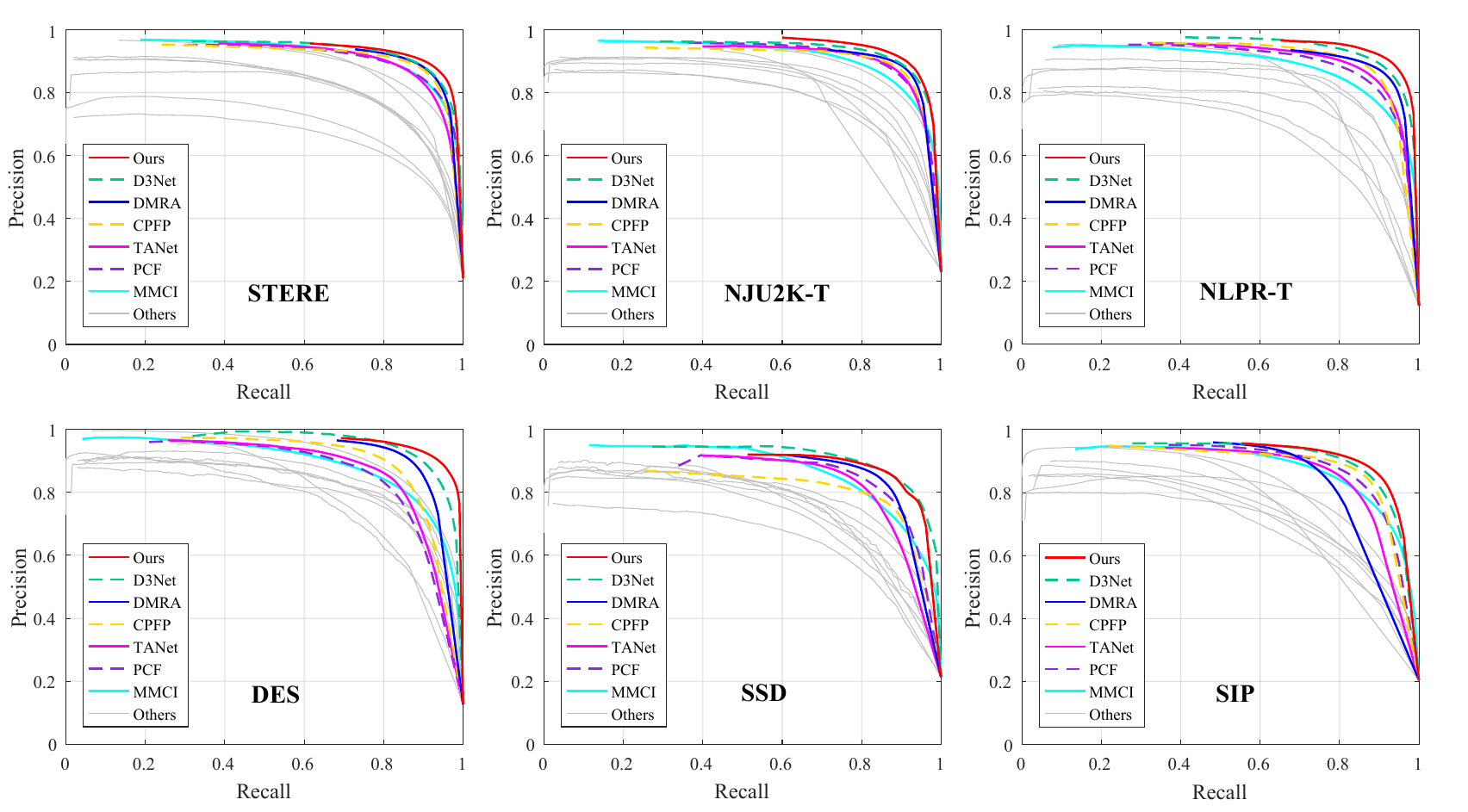}
		\end{overpic}\vspace{-0.3cm}
		\caption{Quantitative comparisons of our MCI-Net with 14 SOTA methods on six challenging benchmark datasets.}
		\label{fig:pr}
	\end{figure*}

	\subsection{Comparison With State-of-the-Arts}
	We compare our model with 14 state-of-the-art RGB-D based SOD methods, including five classical non-deep methods: ACSD \cite{ju2014depth}, LBE \cite{feng2016local}, DCMC \cite{cong2016saliency}, MDSF \cite{song2017depth}, SE \cite{guo2016salient}, and nine CNN-based methods: DF \cite{qu2017rgbd}, AFNet \cite{wang2019adaptive}, CTMF \cite{han2017cnns}, MMCI \cite{chen2019multi}, PCF \cite{chen2018progressively}, TANet \cite{chen2019three}, CPFP \cite{zhao2019contrast}, DMRA \cite{piao2019depth} and D$^{3}$Net \cite{fan2019rethinking}. Note that all the saliency maps of the above methods are provided by the authors and we evaluate them with the same settings.


	\textbf{Quantitative Evaluation.} As shown in \tabref{tab:3}, our method achieves the best scores on five datasets with respect to four metrics, compared with its counterparts. According to the average ranking $(AR)$ in \tabref{tab:3}, the overall ranking of the proposed \ourmodel~ is the highest. In addition, \figref{fig:pr} shows the overall evaluation results of the \emph{PR} curves of our method and comparative methods on six challenging benchmark datasets. Note that the seven best methods are shown in color, while the other methods are in light gray.
	\begin{table}[thp!]
		\small
		\centering
		\renewcommand{\tabcolsep}{0.9mm}
		
		\caption{Benchmarking results of five traditional methods and nine deep learning-based methods on six RGB-D saliency datasets. Here, we adopt mean $ F_{\beta} $ and mean $ E_{\phi} $ \cite{fan2018enhanced}. The best three results are highlighted in {\color{red}{red}}, {\color{blue}{blue}} and {\color{green}{green}}. $\uparrow$ \& $\downarrow$ denote larger and smaller is better, respectively. $ \dag $ denotes the CNN-based RGB-D methods. \emph{AR} denotes the average ranking of each method. }\label{tab:3}
		\begin{center}
			\begin{tabular*}{1\linewidth}
				{rlc|ccccc|cccc} 
				\toprule
				&&\multirow{2}{*}{Metric}&ACSD&LBE&DCMC& MDSF&SE&DF$^\dag $&AFNet$ ^\dag $&CTMF$ ^\dag $ &MMCI$ ^\dag $ \\
				& & &  \cite{ju2014depth} & \cite{feng2016local} &\cite{cong2016saliency} & \cite{song2017depth} &\cite{guo2016salient} &  \cite{qu2017rgbd} &\cite{wang2019adaptive} &\cite{han2017cnns} &\cite{chen2019multi} \\
				\midrule
				\multirow{4}{*}{\rotatebox{90}{\emph{STERE}}} &\multirow{4}{*}{\rotatebox{90}{\cite{niu2012leveraging}}} & $\mathcal{M}\downarrow $& 
				0.200  & 0.250  & 0.148  & 0.176  & 0.143  & 0.141  & 0.075  & 0.086  & 0.068  \\
				& & $ S_{\alpha}\uparrow $ & 0.692  & 0.660  & 0.731  & 0.728  & 0.708  & 0.757  & 0.825  & 0.848  & 0.873  \\
				& & $ F_{\beta}\uparrow $ & 0.478  & 0.501  & 0.590  & 0.527  & 0.610  & 0.617  & 0.806  & 0.758  & 0.813  \\
				& & $ E_{\phi}\uparrow $ & 0.592  & 0.601  & 0.655  & 0.614  & 0.665  & 0.691  & 0.872  & 0.841  & 0.873 \\
				\midrule
				\multirow{4}{*}{\rotatebox{90}{\emph{NJU2K-T}}} &\multirow{4}{*}{\rotatebox{90}{\cite{ju2014depth}}} & $\mathcal{M}\downarrow $& 0.202  & 0.153  & 0.172  & 0.157  & 0.169  & 0.141  & 0.100  & 0.085  & 0.079  \\
				& & $ S_{\alpha}\uparrow $ & 0.699  & 0.695  & 0.686  & 0.748  & 0.664  & 0.763  & 0.772  & 0.849  & 0.858  \\
				& & $ F_{\beta}\uparrow $ & 0.512  & 0.606  & 0.556  & 0.628  & 0.583  & 0.650  & 0.764  & 0.779  & 0.793  \\
				& & $ E_{\phi}\uparrow $ & 0.593  & 0.655  & 0.619  & 0.677  & 0.624  & 0.696  & 0.826  & 0.846  & 0.851  \\
				\midrule
				\multirow{4}{*}{\rotatebox{90}{\emph{NLPR-T}}} &\multirow{4}{*}{\rotatebox{90}{\cite{peng2014rgbd}}}& $\mathcal{M}\downarrow $& 0.179  & 0.081  & 0.117  & 0.095  & 0.091  & 0.085  & 0.058  & 0.056  & 0.059   \\
				& & $ S_{\alpha}\uparrow $ & 0.673  & 0.762  & 0.724  & 0.805  & 0.756  & 0.802  & 0.799  & 0.860  & 0.856    \\
				& & $ F_{\beta}\uparrow $ & 0.429  & 0.736  & 0.543  & 0.649  & 0.624  & 0.664  & 0.755  & 0.740  & 0.737   \\
				& & $ E_{\phi}\uparrow $ & 0.578  & 0.719  & 0.684  & 0.745  & 0.742  & 0.755  & 0.851  & 0.840  & 0.841 \\
				\midrule
				\multirow{4}{*}{\rotatebox{90}{\emph{DES}}}&	\multirow{4}{*}{\rotatebox{90}{\cite{cheng2014depth}}}& $\mathcal{M}\downarrow $& 0.169  & 0.208  & 0.111  & 0.122  & 0.090  & 0.093  & 0.068  & 0.055  & 0.065  \\
				& & $ S_{\alpha}\uparrow $ & 0.728  & 0.703  & 0.707  & 0.741  & 0.741  & 0.752  & 0.770  & 0.863  & 0.848 \\
				& & $ F_{\beta}\uparrow $ & 0.513  & 0.576  & 0.542  & 0.523  & 0.617  & 0.604  & 0.713  & 0.756  & 0.735 \\
				& & $ E_{\phi}\uparrow $ & 0.612  & 0.649  & 0.632  & 0.621  & 0.707  & 0.684  & 0.810  & 0.826  & 0.825  \\
				\midrule
				\multirow{4}{*}{\rotatebox{90}{\emph{SSD}}} &\multirow{4}{*}{\rotatebox{90}{\cite{zhu2017three}}} & $\mathcal{M}\downarrow $& 0.203  & 0.278  & 0.169  & 0.192  & 0.165  & 0.142  & 0.118  & 0.099  & 0.082  \\
				& & $ S_{\alpha}\uparrow $ & 0.675  & 0.621  & 0.704  & 0.673  & 0.675  & 0.747  & 0.714  & 0.776  & 0.813 \\
				& & $ F_{\beta}\uparrow $ & 0.469  & 0.489  & 0.572  & 0.470  & 0.564  & 0.624  & 0.672  & 0.689  & 0.721 \\
				& & $ E_{\phi}\uparrow $ & 0.566  & 0.574  & 0.646  & 0.576  & 0.631  & 0.690  & 0.762  & 0.796  & 0.796 \\
				\midrule
				\multirow{4}{*}{\rotatebox{90}{\emph{SIP}}} &\multirow{4}{*}{\rotatebox{90}{\cite{fan2019rethinking}}} & $\mathcal{M}\downarrow $& 0.172  & 0.200  & 0.186  & 0.167  & 0.164  & 0.185  & 0.118  & 0.139  & 0.086  \\
				& & $ S_{\alpha}\uparrow $ & 0.732  & 0.727  & 0.683  & 0.717  & 0.628  & 0.653  & 0.720  & 0.716  & 0.833  \\
				& & $ F_{\beta}\uparrow $ & 0.542  & 0.571  & 0.499  & 0.568  & 0.515  & 0.464  & 0.702  & 0.608  & 0.771 \\
				& & $ E_{\phi}\uparrow $ & 0.614  & 0.651  & 0.598  & 0.645  & 0.592  & 0.565  & 0.793  & 0.705  & 0.845 \\
				\midrule
				\multicolumn{3}{c|}{\emph{AR}} &  13.83  & 12.75  & 12.92  & 12.13  & 12.08  & 10.88  & 8.71  & 8.08  & 7.17   \\
				
				\bottomrule	
			\end{tabular*}
		\end{center}
	\end{table}
	
	\begin{table}[thp!]
		\small
		\centering
		\renewcommand{\arraystretch}{0.9}
		\renewcommand{\tabcolsep}{0.9mm}
		
		\caption{Continuation of Tab. 2.}\label{tab:continuation3}
		\begin{center}
			
			\begin{tabular*}{0.8\linewidth}
				{rlc|ccccc|ccccccccc|c} 
				\toprule
				&&\multirow{2}{*}{Metric} &PCF$ ^\dag $ &TANet$ ^\dag $&CPFP$ ^\dag $&DMRA$ ^\dag $&D$^{3}$Net$ ^\dag $&\textbf{MCI-Net}\\
				& &  & \cite{chen2018progressively}& \cite{chen2019three}& \cite{zhao2019contrast}& \cite{piao2019depth}& \cite{fan2019rethinking}&Ours\\
				\midrule
				\multirow{4}{*}{\rotatebox{90}{\emph{STERE}}} &\multirow{4}{*}{\rotatebox{90}{\cite{niu2012leveraging}}} & $\mathcal{M}\downarrow $& 0.064  & 0.060  &\color{green}{0.051}  & \color{blue}{0.047}  & 0.054  & \color{red}{0.042}\\
				& & $ S_{\alpha}\uparrow $ & 0.875  & 0.871  & 0.879  &\color{green}{0.886}   &\color{blue}{0.891}& \color{red}{0.901}\\
				& & $ F_{\beta}\uparrow $ & 0.818  & 0.828  & 0.841  & \color{blue}{0.868}  &\color{green}{0.844}& \color{red}{0.872} \\
				& & $ E_{\phi}\uparrow $ & 0.887  & 0.893  & \color{green}{0.912}& \color{blue}{0.920} & 0.908  & \color{red}{0.929} \\
				\midrule
				\multirow{4}{*}{\rotatebox{90}{\emph{NJU2K-T}}} &\multirow{4}{*}{\rotatebox{90}{\cite{ju2014depth}}} & $\mathcal{M}\downarrow $& 0.059  & 0.060  &\color{green}{0.053}   & \color{blue}{0.051}  & \color{blue}{0.051}  & \color{red}{0.050} \\
				& & $ S_{\alpha}\uparrow $& 0.877  & 0.878  & 0.878  & \color{green}{0.886}  &\color{blue}{0.895}   & \color{red}{0.900} \\
				& & $ F_{\beta}\uparrow $& 0.840  & 0.841  &\color{green}{0.850}  &\color{red}{0.873}   & \color{blue}{0.860}  & \color{red}{0.873} \\
				& & $ E_{\phi}\uparrow $& 0.895  & 0.895  &\color{green}{0.910}  &\color{red}{0.920} & \color{blue}{0.912} & \color{red}{0.920} \\
				\midrule
				\multirow{4}{*}{\rotatebox{90}{\emph{NLPR-T}}} &\multirow{4}{*}{\rotatebox{90}{\cite{peng2014rgbd}}}& $\mathcal{M}\downarrow $& 0.044  & 0.041  & 0.036  &\color{blue}{0.031}  & \color{green}{0.034} & \color{red}{0.027} \\
				& & $ S_{\alpha}\uparrow $& 0.874  & 0.886  & 0.888  & \color{green}{0.899}  & \color{blue}{0.906}  &\color{red}{0.917}  \\
				& & $ F_{\beta}\uparrow $ & 0.802  & 0.819  & 0.840  & \color{blue}{0.865} & \color{green}{0.853}  & \color{red}{0.890} \\
				& & $ E_{\phi}\uparrow $& 0.887  & 0.902  & 0.918  & \color{blue}{0.940} &\color{green}{0.923}  & \color{red}{0.947}\\
				\midrule
				\multirow{4}{*}{\rotatebox{90}{\emph{DES}}}&	\multirow{4}{*}{\rotatebox{90}{\cite{cheng2014depth}}}& $\mathcal{M}\downarrow $& 0.049  & 0.046  &\color{green}{0.038}  & \color{blue}{0.030}  & \color{blue}{0.030}  & \color{red}{0.024} \\
				& & $ S_{\alpha}\uparrow $& 0.842  & 0.858  & 0.872  & \color{green}{0.900}  & \color{blue}{0.904}  & \color{red}{0.927} \\
				& & $ F_{\beta}\uparrow $& 0.765  & 0.790  & 0.824  &\color{blue}{0.873}  & \color{green}{0.859} & \color{red}{0.897}\\
				& & $ E_{\phi}\uparrow $& 0.838  & 0.863  & 0.889  & \color{blue}{0.933}  & \color{green}{0.909}  & \color{red}{0.957}\\
				\midrule
				\multirow{4}{*}{\rotatebox{90}{\emph{SSD}}} &\multirow{4}{*}{\rotatebox{90}{\cite{zhu2017three}}} & $\mathcal{M}\downarrow $ &\color{green}{0.062} & 0.063  & 0.082  & \color{blue}{0.058}  &\color{blue}{0.058}   & \color{red}{0.052}\\
				& & $ S_{\alpha}\uparrow $& 0.841  & 0.839  & 0.807  & \color{green}{0.857} & \color{red}{0.866} &\color{blue}{0.860} \\
				& & $ F_{\beta}\uparrow $& 0.777  & 0.773  & 0.747  & \color{red}{0.828}   &\color{green}{0.818}   &\color{blue}{0.820} \\
				& & $ E_{\phi}\uparrow $& 0.856  & 0.861  & 0.839  &\color{blue}{0.897}  &\color{green}{0.887}   &\color{red}{0.901}  \\
				\midrule
				\multirow{4}{*}{\rotatebox{90}{\emph{SIP}}} &\multirow{4}{*}{\rotatebox{90}{\cite{fan2019rethinking}}} & $\mathcal{M}\downarrow $& 0.071  & 0.075  & \color{green}{0.064}  & 0.085  &\color{blue}{0.063}  &\color{red}{0.056}  \\
				& & $ S_{\alpha}\uparrow $ & 0.842  & 0.835  &\color{green}{0.850}  & 0.806  & \color{blue}{0.864}  & \color{red}{0.867} \\
				& & $ F_{\beta}\uparrow $ & 0.814  & 0.803  &\color{green}{0.821} & 0.811  & \color{blue}{0.832} & \color{red}{0.840} \\
				& & $ E_{\phi}\uparrow $ & 0.878  & 0.870  &\color{green}{0.893}  & 0.844  &\color{blue}{0.894}   &\color{red}{0.909} \\
				\midrule
				\multicolumn{3}{c|}{\emph{AR}}& 5.33  & 5.13  & 4.13  & \color{green}{2.79}  & \color{blue}{2.54} & \color{red}{1.08} \\
				
				\bottomrule	
			\end{tabular*}
		\end{center}
	\end{table}
	
	\begin{figure*}[t]
		\tiny
		\centering
		\begin{overpic}[width=\linewidth]{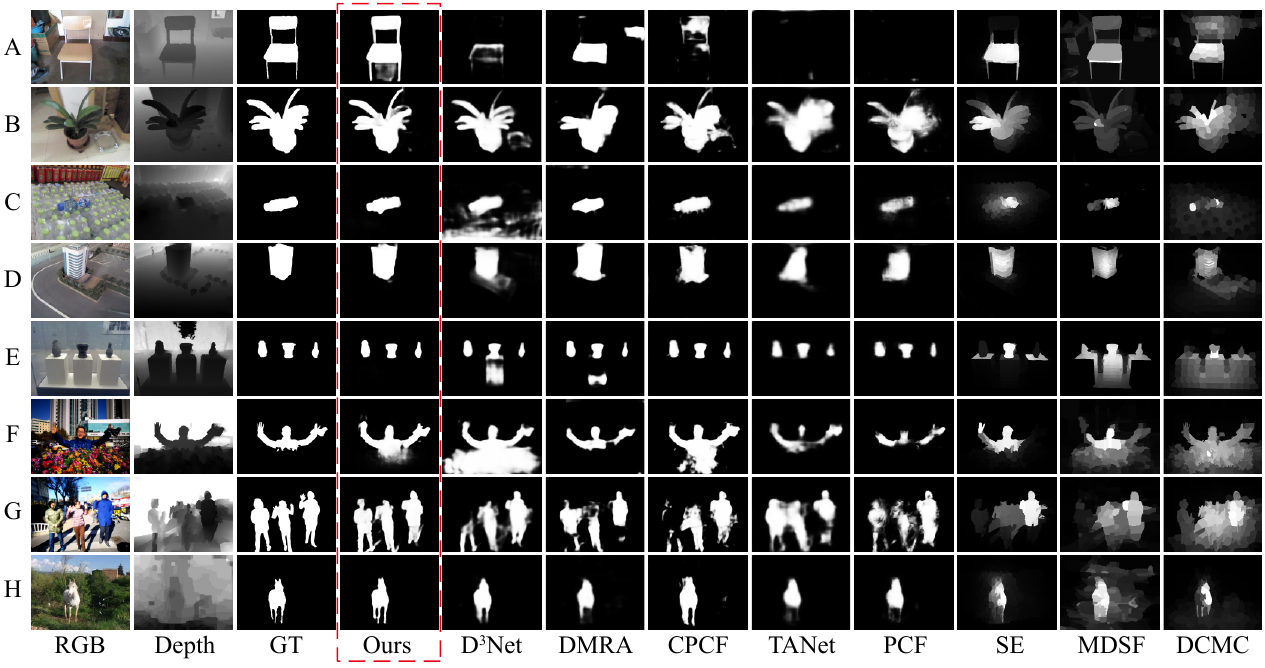}
			
		\end{overpic}
		\caption{Visual comparisons to top nine SOTA methods under different challenging situations.}
		\label{fig:vc}
	\end{figure*}
	
	\textbf{Qualitative Comparisons.} We visually compare \ourmodel~ with other SOTA methods, as shown in \figref{fig:vc}. From these results, it can be seen that the saliency map of our method is closer to the ground truth. Further, it can be also observed that the proposed method completely highlights the salient object regions, and excels in dealing with various challenging scenarios, including complex backgrounds (the $A$ and $F$ rows), transparent objects ($C$ row), low- quality depth maps (the $B$  and $H$ rows), and multiple and small objects (the $E$  and $G$ rows). The visual comparison further validates the effectiveness and robustness of our \ourmodel.
	
\begin{figure*}
		\centering
		\begin{overpic}[width=\linewidth]{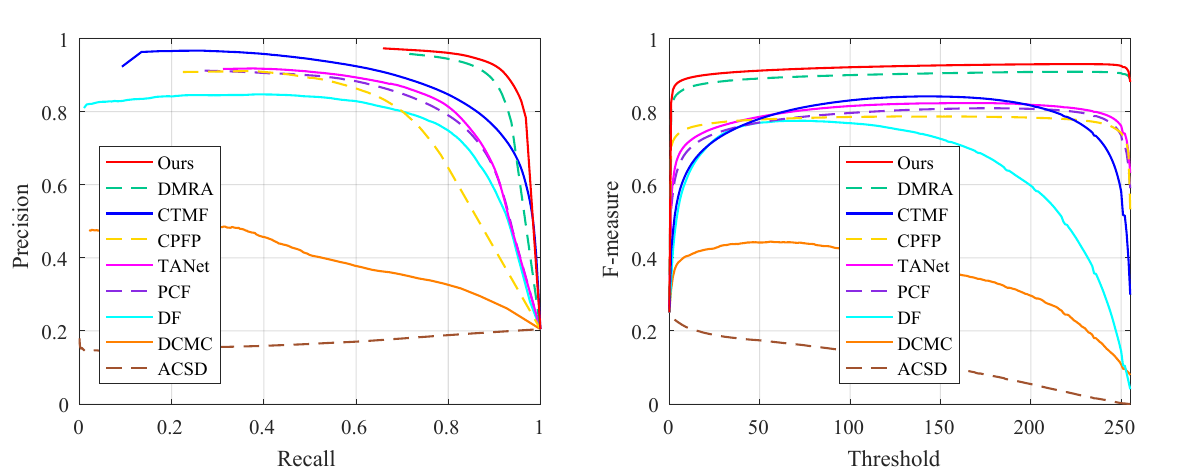}
		\end{overpic}\vspace{-0.3cm}
		\caption{\emph{PR} curves and Threshold-F-measure curves (from left to right) of
			different models on DUTLF \cite{piao2019depth}.}
		\label{fig:DUTLF}
	\end{figure*}

\begin{figure*}[thp!]
		\centering
		\begin{overpic}[width=\linewidth]{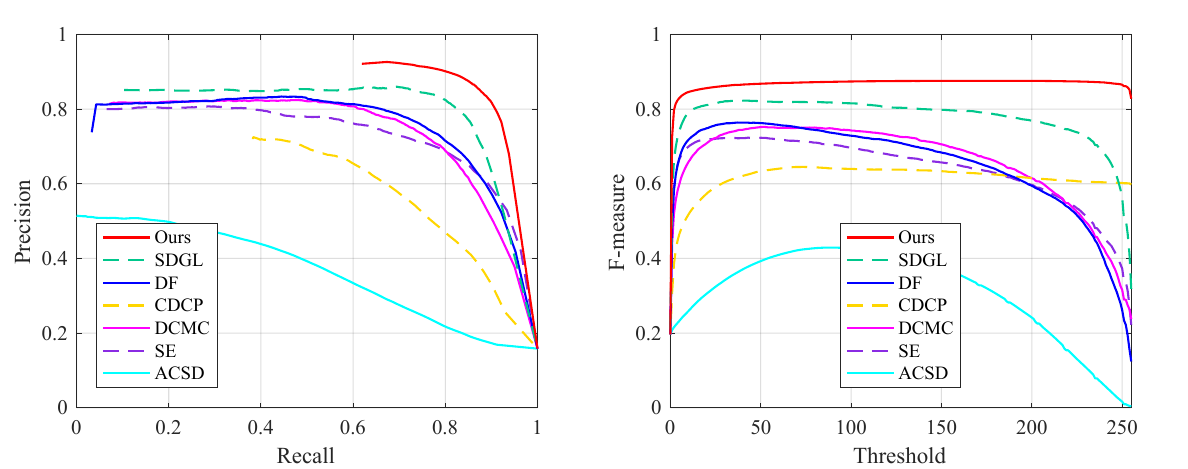}
		\end{overpic}
		\caption{\emph{PR} curves and Threshold-F-measure curves (from left to right) of our and other models on VT1000 \cite{tu2019rgb}.}
		\label{fig:VT1000}
	\end{figure*}

\begin{figure*}[thp!]
		\centering
		\begin{overpic}[width=\linewidth]{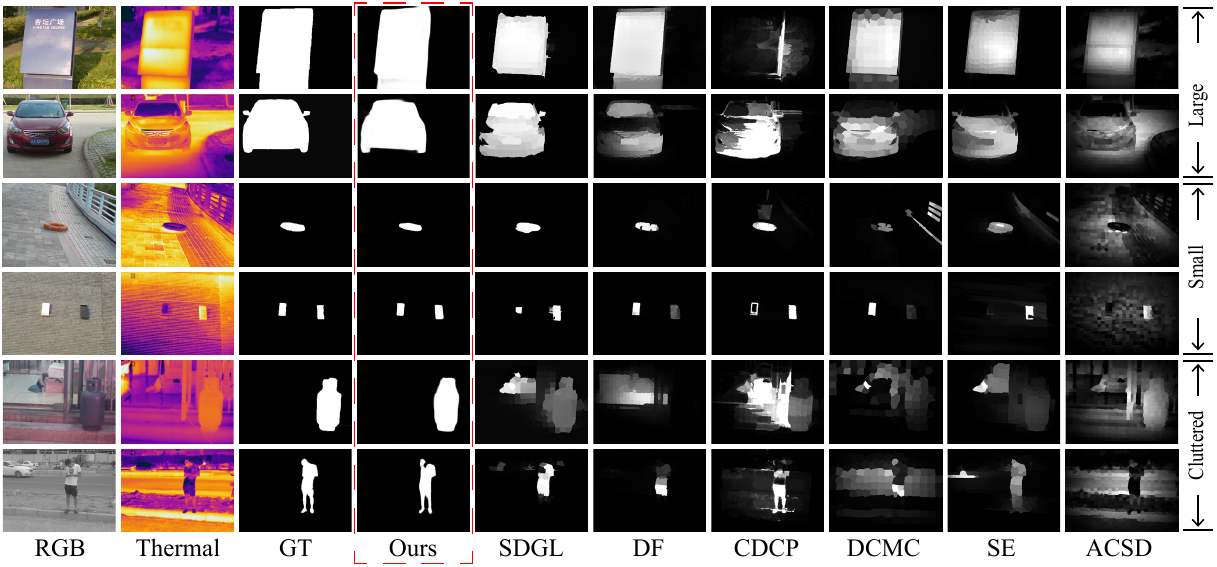}
		\end{overpic}
		\caption{Visual comparison of typical scenes in VT1000 datasets.}
		\label{fig:VC-VT1000}
	\end{figure*}

	In the supplementary experiments, for DUTLF \cite{piao2019depth}, \tabref{tab:DUTLF} and \figref{fig:DUTLF} show the quantitative comparison results between our model and several SOTA methods including DMRA \cite{piao2019depth}, CTMF \cite{han2017cnns}, CPFP \cite{zhao2019contrast}, TANet \cite{chen2019three}, PCF \cite{chen2018progressively}, DF \cite{qu2017rgbd}, DCMC \cite{cong2016saliency}, and ACSD \cite{ju2014depth}. For VT1000 \cite{tu2019rgb}, we compare the proposed model with the SOTA models including SDGL \cite{tu2019rgb}, DF \cite{qu2017rgbd}, CDCP \cite{zhu2017innovative} DCMC \cite{cong2016saliency}, SE \cite{guo2016salient}, and ACSD \cite{ju2014depth}. The quantitative experimental results are shown in \tabref{tab:VT1000} and \figref{fig:VT1000}. The visual comparison of some challenging scenes is shown in \figref{fig:VC-VT1000}, which mainly includes typical scenes with small-size, large-size objects and cluttered backgrounds. From these results, these can be observed that our model still performs better than all comparison methods and also has great potential in multi-modal SOD.
	
	\begin{table}[thp!]
		\centering
		\renewcommand{\arraystretch}{1.2}
		\renewcommand{\tabcolsep}{0.5mm}
		
		\caption{Quantitative evaluation on the DUTLF \cite{piao2019depth} dataset. The best results are highlighted in \textbf{bold}.   Here, we adopt mean $ F_{\beta} $ and mean $ E_{\phi} $ \cite{fan2018enhanced}. $\uparrow$ \& $\downarrow$ denote larger and smaller is better, respectively. $ \dag $ denotes the CNN-based RGB-D methods.}\label{tab:DUTLF}
		
		\begin{tabular*}{1\linewidth}{c|cc|cccccc|c} 
			\toprule
			\multirow{2}{*}{Metric}& ACSD& DCMC & DF$^\dag $ &PCF$^\dag $ & TANet$ ^\dag $&CPFP$ ^\dag $&CTMF$ ^\dag $&DMRA$ ^\dag $&\textbf{MCI-Net}\\
			&\cite{ju2014depth} & \cite{cong2016saliency}&\cite{qu2017rgbd} &\cite{chen2018progressively} &\cite{chen2019three}&\cite{zhao2019contrast}&\cite{han2017cnns}&\cite{piao2019depth}&Ours\\
			
			\midrule
			$\mathcal{M}\downarrow $&0.332&0.243&0.145&0.100&0.093&0.099&0.097&0.048&\textbf{0.039}\\
			$ S_{\alpha}\uparrow $ &0.361&0.499&0.730&0.801&0.808&0.749&0.831&0.889&\textbf{0.906}\\
			$ F_{\beta}\uparrow $ &0.106&0.318&0.585&0.741&0.761&0.696&0.747&0.885&\textbf{0.902}\\
			$ E_{\phi}\uparrow $ &0.432&0.540&0.665&0.821&0.831&0.760&0.810&0.927&\textbf{0.939}\\
			
			\bottomrule	
		\end{tabular*}
	\end{table}
	
	\begin{table}[thp!]
		\centering
		\renewcommand{\arraystretch}{1.2}
		\renewcommand{\tabcolsep}{1.5mm}
		
		\caption{Quantitative evaluation on the VT1000 \cite{tu2019rgb} dataset, in which the mean $ F_{\beta} $ and mean $ E_{\phi}$ are used. $ \dag $ denotes the CNN-based method.}\label{tab:VT1000}
		\begin{center}

			\begin{tabular*}{0.88\linewidth}{c|cccccc|c} 
				\toprule
				\multirow{2}{*}{Metric}&ACSD&SE &DCMC &CDCP &DF$^ \dag $ & SDGL &\textbf{MCI-Net}\\
				
				&\cite{ju2014depth} & \cite{guo2016salient} & \cite{cong2016saliency} &\cite{zhu2017innovative} &\cite{qu2017rgbd}&\cite{tu2019rgb}&Ours\\
				
				\midrule
				$\mathcal{M}\downarrow $&0.223&0.121&0.116&0.137&0.116&0.099&\textbf{0.040}\\
				$ S_{\alpha}\uparrow $ &0.537&0.684&0.717&0.655&0.703&0.083&\textbf{0.872}\\
				$ F_{\beta}\uparrow $ &0.279&0.569&0.600&0.534&0.717&0.788&\textbf{0.837}\\
				$ E_{\phi}\uparrow $ &0.513&0.668&0.692&0.687&0.670&0.795&\textbf{0.899}\\
				
				\bottomrule	
			\end{tabular*}
		\end{center}
	\end{table}

	\subsection{Ablation Study}
	In order to verify the relative importance and specific contribution of each component of our proposed model, we use the same network settings as above for the ablation study.
	
	\textbf{A. HHA-E.} To explore the contribution of coding to HHA-E through the contrast enhancement method, we use RGB images and original depth images (A1) or RGB images and depth maps directly encoded with HHA (A2) as different input image pairs for testing. Columns A1 and A2 in \tabref{tab:4} show that HHA-E promotes performance improvement.
	
	\textbf{B. CMRM} \emph{vs.} \textbf{MLFM.} The effective fusion of cross-modal and multi-level features is the key to integrating multi-source features. In order to illustrate the effectiveness of the two proposed modules, we use a general fusion method (\emph{i.e.}, directly superimpose two features) to replace the CMRM unit (B1) and MLFM unit (B2), respectively. From \tabref{tab:4}, we observe that CMRM and MLFM increase the performance of the network in four metrics. This suggests that the two proposed modules can help our network more accurately distinguish the salient regions.
	
	\textbf{C. Multi-level Interactive Integration Strategy.} To verify the effectiveness of the proposed multi-level integration strategy, we first conduct a comparison experiment by removing the FIM structure (C1). Then, in the top-down process, we reshape the multi-level features to the same size, and use a simple fusion strategy to directly connect the corresponding features (C2). Finally, in the process from right to left, we fuse two adjacent features of different levels and input them to the next layer of the pyramid (C3). The performance improvements are shown \tabref{tab:4}, demonstrating the importance of the multi-level integration strategy for \ourmodel.
	
	\textbf{D. EPA Loss.} We utilize the EPA loss to make the network pay more attention to the overall structure of the image and mitigate the impact caused by the uneven distribution of features. Further, we also experimentally tested the effects of different loss functions, including BCE (D1), wIoU (D2), and  combining the BCE and wIoU loss (D3). From the comparison results in \tabref{tab:4}, it can be observed that the EAP loss improves the performance of our \ourmodel.

	\begin{table}[t!]
		
		\centering
		\renewcommand{\arraystretch}{1.2}
		\renewcommand{\tabcolsep}{0.7mm}
		
		\caption{Ablation study on RGB-D saliency datasets. The best result in each row is highlighted in \textbf{bold}.}
		\label{tab:4}
		
		\begin{tabular*}{1\linewidth}{cr|c|cc|cc|ccc|ccc} 
			\toprule
			& Metric & Ours & A1 & A2 & B1 & B2 & C1 & C2 & C3 & D1 & D2 & D3 \\
			
			\midrule
			\multirow{4}{*}{\rotatebox{90}{STERE \cite{niu2012leveraging}}}
			& $\mathcal{M}\downarrow $ &\textbf{0.042} & 0.050  & 0.048  & 0.061  & 0.068  & 0.056  & 0.059  & 0.059  & 0.056  & 0.048  & 0.045 \\
			&  $ S_{\alpha}\uparrow $  &\textbf{0.901} & 0.892  & 0.896  & 0.867  & 0.860  & 0.871  & 0.868  & 0.871  & 0.872  & 0.885  & 0.894 \\
			&  $ F_{\beta}\uparrow $  &\textbf{0.872} &  0.862  & 0.866  & 0.829  & 0.822  & 0.838  & 0.832  & 0.833  & 0.838  & 0.856  & 0.868  \\
			&  $ E_{\phi}\uparrow $  & \textbf{0.929}&  0.918  & 0.923  & 0.901  & 0.894  & 0.906  & 0.904  & 0.903  & 0.909  & 0.918  & 0.924  \\
			\midrule
			\multirow{4}{*}{\rotatebox{90}{SSD \cite{zhu2017three}}} 
			& $\mathcal{M}\downarrow $ &\textbf{0.052} & 0.061  & 0.057  & 0.082  & 0.088  & 0.070  & 0.075  & 0.077  & 0.062  & 0.056  & 0.054  \\
			& $ S_{\alpha}\uparrow $&\textbf{0.860}&  0.850  & 0.854  & 0.816  & 0.812  & 0.827  & 0.829  & 0.823  & 0.850  & 0.854  & 0.857 \\
			& $ F_{\beta}\uparrow $   &\textbf{0.820}& 0.811  & 0.815  & 0.783  & 0.773  & 0.791  & 0.797  & 0.787  & 0.787  & 0.790  & 0.895 \\
			& $ E_{\phi}\uparrow $  &\textbf{0.901}& 0.899  & 0.895  & 0.855  & 0.848  & 0.864  & 0.866  & 0.861  & 0.886  & 0.894  & 0.897  \\
			\midrule
			\multirow{4}{*}{\rotatebox{90}{SIP \cite{fan2019rethinking}}} 
			& $\mathcal{M}\downarrow $ &\textbf{0.056}& 0.063  & 0.059  & 0.076  & 0.072  & 0.072  & 0.072  & 0.075  & 0.066  & 0.061  & 0.059 \\
			& $ S_{\alpha}\uparrow $ &\textbf{0.867} & 0.856  & 0.860  & 0.844  & 0.834  & 0.841  & 0.849  & 0.848  & 0.854 & 0.859  & 0.863  \\
			& $ F_{\beta}\uparrow $ &\textbf{0.840} & 0.838  & 0.839  & 0.813  & 0.806  & 0.825  & 0.823  & 0.817  & 0.821  & 0.829  & 0.835\\
			& $ E_{\phi}\uparrow $ &\textbf{0.909}& 0.902  & 0.905  & 0.867  & 0.861  & 0.877  & 0.872  & 0.871  & 0.889  & 0.893  & 0.897 \\
			\bottomrule	
		\end{tabular*}
	\end{table}
	
	\section{Conclusion}
	
	
	In this paper, we propose a new SOD framework for RGB-D, termed \ourmodel. Our \ourmodel~ includes two key components: a cross-modal feature learning network and a multi-level interactive integration network. The cross-modal feature learning network is used to learn high-level features for RGB images and depth cues, effectively fusing the two sources and exploiting their correlations. The multi-level interactive integration network can fuse the features of each level through a bottom-up strategy in a pyramid style, which also propagates the features of the last convolutional layer back to the previous layers to reduce the effect of noise and information loss. Experimental results on six challenging datasets demonstrate that our \ourmodel~ outperforms 14 SOTA methods, and the comprehensive ablation study also validates the effectiveness of all key components.

	In future work, the importance and effectiveness of cross-modal features features might be worth exploring. In addition, with the development of monocular depth estimation technology over the past few years \cite{fu2018deep,yan2018monocular,godard2019digging}, we will extend our model to other saliency related tasks, such as V-SOD \cite{fan2019shifting}, Co-SOD \cite{fan2020taking}, and camouflaged object detection \cite{fan2020camouflaged}.

	\section*{Acknowledgments}
	This research was supported by a grant from the Sichuan Major Science and Technology Special Foundation (No.2018GZDZX0017).
	\bibliographystyle{elsarticle-num} 
	\bibliography{NC-MCINet.bib}
\end{document}